\documentclass{ecai} 

\usepackage{latexsym}
\usepackage{amssymb}
\usepackage{amsmath}
\usepackage{amsthm}
\usepackage{booktabs}
\usepackage{enumitem}
\usepackage{graphicx}
\usepackage{color}
\usepackage[table]{xcolor}

\newcommand{\BibTeX}{B\kern-.05em{\sc i\kern-.025em b}\kern-.08em\TeX}

\begin{document}


\begin{frontmatter}


\paperid{1780} 


\title{Corner2Net: Detecting Objects as Cascade Corners}


\author[A]{\fnms{Chenglong}~\snm{Liu}\footnote[\dagger]{Equal contribution.  Email: liuchenglong20@mails.ucas.ac.cn}}
\author[A]{\fnms{Jintao}~\snm{Liu}\footnotemark[\dagger]}
\author[A]{\fnms{Haorao}~\snm{Wei}} 
\author[A]{\fnms{Jinze}~\snm{Yang}} 
\author[A]{\fnms{Liangyu}~\snm{Xu}} 
\author[B]{\fnms{Yuchen}~\snm{Guo}\thanks{Corresponding Author.}} 
\author[B]{\fnms{Lu}~\snm{Fang}} 

\address[A]{University of Chinese Academy of Sciences}
\address[B]{Beijing National Research Center for Information Science and Technology, Tsinghua University, Beijing, China}


\begin{abstract}
The corner-based detection paradigm enjoys the potential to produce high-quality boxes. But the development is constrained by three factors: 1) Hard to match corners. Heuristic corner matching algorithms can lead to incorrect boxes, especially when similar-looking objects co-occur. 2) Poor instance context. Two separate corners preserve few instance semantics, so it is difficult to guarantee getting both two class-specific corners on the same heatmap channel. 3) Unfriendly backbone. The training cost of the hourglass network is high. Accordingly, we build a novel corner-based framework, named Corner2Net. To achieve the corner-matching-free manner, we devise the cascade corner pipeline which progressively predicts the associated corner pair in two steps instead of synchronously searching two independent corners via parallel heads. Corner2Net decouples corner localization and object classification. Both two corners are class-agnostic and the instance-specific bottom-right corner further simplifies its search space. Meanwhile, RoI features with rich semantics are extracted for classification. Popular backbones (\textit{e.g.}, ResNeXt) can be easily connected to Corner2Net. Experimental results on COCO show Corner2Net surpasses all existing corner-based detectors by a large margin in accuracy and speed. 
\end{abstract}

\end{frontmatter}


\section{Introduction} \label{section:intro}
\begin{figure}[!t]
\includegraphics[width=\linewidth]{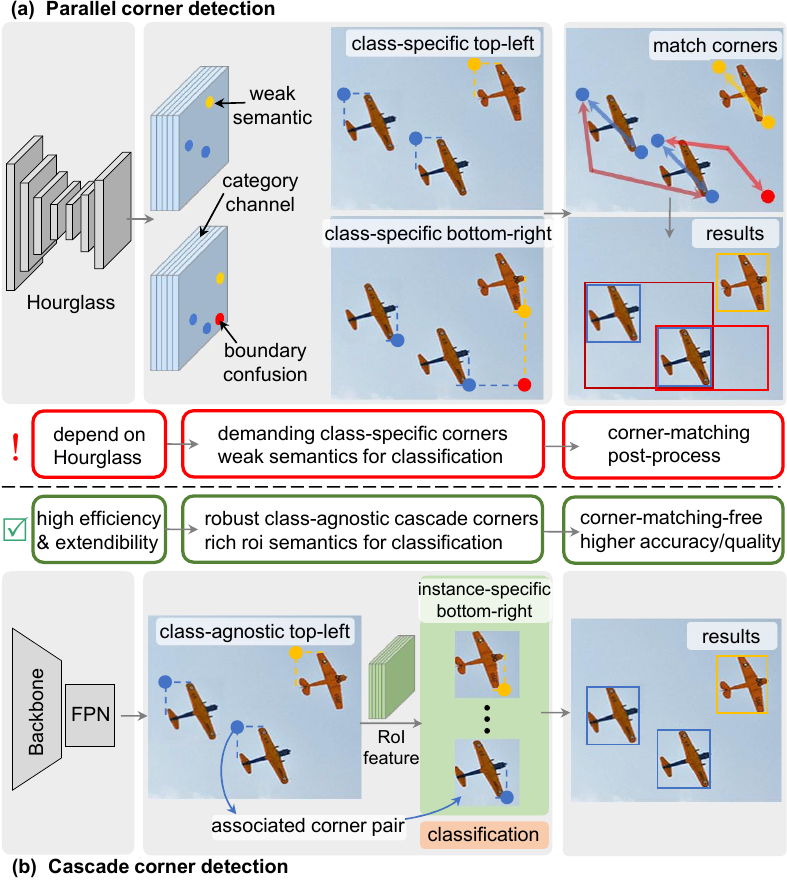}
\caption{Comparision of parallel corner detection and the proposed cascade corner detection. All existing corner-based methods fall into the parallel detection pipeline, which predicts two separate class-specific corners and relies on the corner-matching algorithm when decoding boxes. Corner2Net adopts the proposed cascade corner detection pipeline that decouples the corner localization and object classification. Corner2Net mines all objects via the class-agnostic associated corner pair which is more robust, and the instance-specific bottom-right corner further simplifies its search space. The object category is predicted using RoI features with rich instance semantics. Our Corner2Net runs in a corner-matching-free manner, and it can achieve higher accuracy and efficiency under various backbones and is not limited to the Hourglass network.}
\label{fig:intro}
\vspace{20pt}
\end{figure}

Object detection~\cite{ren2015fasterrcnn,lin2017focalloss,tian2019fcos,redmon2016yolo,liu2023s2cgnet,wei2022humanliker} is a research hotspot in computer vision, which aims to know where and what objects are in a given image. With the development of deep learning~\cite{krizhevsky2017imagenet}, the accuracy and efficiency of object detection have been greatly improved, enabling it to develop into practical applications. 

In the current era, according to the representations of modeling objects, more popular object detectors are the center-based type. They encode a bounding box by taking the object center as a reference and regressing the height/width. More specifically, the classic center-based methods~\cite{ren2015fasterrcnn,cai2018cascadercnn} initialize each pixel as a center point and place a large number of artificially sized anchors. Later, some excellent center-based detectors~\cite{redmon2018yolov3,lin2017focalloss,tian2019fcos,sun2021sparsercnn} are proposed to optimize the learning of dense anchor queries and make them more sparse and efficient. Different from the aforementioned methods, to get rid of anchors relying on manually set hyperparameters, a corner-based detector~\cite{law2018cornernet} has emerged. This model constructs a detection box by predicting its top-left and bottom-right corner keypoints and matching them into a pair. Afterward, some impressive efforts~\cite{duan2019centernet_duan,ijcai2022p203_CornerAffinity,liu2024gigahumandet} have been made to pursue more robust corner matching algorithms with higher quality and higher reliability.

Actually, the center-based modeling approaches can be harder to determine object boundaries via its center, because they directly optimize the bounding box with four degrees of freedom and most training pipelines are less sensitive to precise object boundaries. Compared to them, the corner-based methods only require two independent corner keypoints with two degrees of freedom each. Further, this corner-based paradigm is similar to what annotators do, which is to label a precise bounding box from the top-left corner to the bottom-right corner. Thus, the idea of detecting an object with two corners implies strong human empirical knowledge. We hold the view that the corner-based paradigm enjoys excellent research value and the potential to produce high-quality boxes. 

In this era where center-based detectors dominate, there are several reasons why corner-based ones are not as popular. First, and most importantly, the process of constructing a box from two independent corners relies on a heuristic matching algorithm, which needs to filter out ambiguous corner keypoints caused by boundary confusion and find out the corner pair belonging to the same object instance (shown in Figure~\ref{fig:intro} (a)), while its reliability and preciseness are not perfect. Second, it is difficult to predict class-specific keypoints because the corners preserve poor instance context. This is unfavorable for the way of obtaining the object category by the same channel ID of the heatmap where the corresponding corner pair lies. Sometimes one of the two class-specific corners is wrongly estimated due to the weak robustness and learning difficulty, which directly leads to the loss of a detection box. Third, the existing corner-based detectors all adopt the keypoint-friendly backbone Hourglass~\cite{newell2016stacked_hourglass} network, which has low training efficiency and slow inference speed, making it difficult to deploy in practical applications.

Accordingly, to address these issues, we build a novel corner-based framework, named Corner2Net, which focuses more on each instance and frees the corner detector from the shackles it carried before. In general, Corner2Net decouples corner localization and object classification and changes the conventional parallel prediction to a cascade prediction pipeline (see Figure~\ref{fig:intro} (b)). To alleviate the learning difficulty and fully mine all the potential objects, Corner2Net locates objects using class-agnostic associated corner pairs instead of hard class-specific independent keypoints. Moreover, each bottom-right corner is determined within the corresponding instance-specific RoI space to further simplify corner search. Obviously, the associated cascade corner pair enables Corner2Net to get rid of the dependence on the matching algorithm and directly produce a box. Meanwhile, to improve classification accuracy, the category of each object is obtained by feeding RoI features into a lightweight head, which maintains rich instance context. For more details, unlike existing corner-based methods~\cite{law2018cornernet,duan2019centernet_duan,ijcai2022p203_CornerAffinity} that use a single feature from Hourglass network, we exploit the multi-level prediction in FPN layers to fully adapt to objects of different sizes when pinpointing corners. In addition, our Corner2Net can easily connect popular backbones (\textit{e.g.}, ResNeXt~\cite{xie2017aggregated_ResNeXt}) and yield a pleasing performance. 

We evaluate the proposed Corner2Net on three challenging datasets, \textit{i.e.}, MS-COCO \cite{lin2014microsoft_coco_dataset}, CityPersons~\cite{cordts2016cityscapes} and UCAS-AOD~\cite{zhu2015orientation}. Corner2Net achieves an AP of 47.8\% under the ResNeXt-101-DCN backbone on the COCO test-dev, exceeding all existing corner-based detectors by a large margin. Without the cumbersome hourglass and corner-matching process, Corner2Net enjoys a 2.1 times faster inference speed than the CornerNet baseline. Furthermore, the remarkable $\mathrm{AP_{80}}$/$\mathrm{AP_{90}}$ of 44.6\%/22.4\% verify the superiority of producing high-quality boxes. 
In addition, on the CityPersons and UCAS-AOD datasets that baseline detector CornerNet struggle with, the proposed Corner2Net gains significant improvements of 36.2\% and 18.0\% on $\mathrm{AP_{50}}$, respectively. This indicates that our Corner2Net enjoys strong robustness and applicability.
We firmly believe that the neat yet efficient Corner2Net can become an excellent baseline in the corner detection paradigm. 

To summarize, our contributions are three folds:
\begin{itemize}
    \item We propose an innovative corner-based detector named Corner2Net, achieving a corner-matching-free manner by modeling objects as associated cascade corners.
    \item Corner-based localization and classification are decoupled and optimized by class-agnostic corner pairs and RoI semantics. Corner2Net can robustly benefit from mainstream backbones.
    \item Corner2Net far outperforms all existing corner detectors in terms of accuracy and speed. Further, the quality of detection boxes is significantly improved.
\end{itemize}

\section{Related Work}
In this section, we briefly review some related approaches, mainly involving center-based and corner-based detectors.
\subsection{Center-based Detectors}

The center-based paradigm is a research hotspot in object detectors. They model the bounding box from the center of the object and encode its length and width. 
The famous Faster-RCNN~\cite{ren2015fasterrcnn} places numerous dense center-based anchors and regresses the offsets between the initial anchors and the ground truths. To improve efficiency and practicability, YOLO series~\cite{redmon2016yolo,redmon2018yolov3} abandons the proposal stage and directly performs regression and classification for each box. 
Libra R-CNN~\cite{pang2019libra_rcnn} investigates the impact of the training process on the anchor-guided detector and optimizes it by the balanced learning at the sample level, feature level, and objective level.
Later, FCOS~\cite{tian2019fcos} performs per-pixel prediction within the center of the object and regresses a 4D vector. 
Later, Dynamic R-CNN~\cite{zhang2020dynamic_rcnn} uses dynamic statistics of proposals instead of static configuration during training to make the detector more effective. Sparse RCNN~\cite{sun2021sparsercnn} discards conventional dense priors and employs a few learnable boxes to detect objects, which eliminates many-to-one label assignment and non-maximum suppression post-processing. 
DETR~\cite{carion2020end_detr} solves the detection problem as a direct set prediction and designs the transformer architecture to query the position of objects based on centers. 
Center-based detectors have undoubtedly achieved great results, however, this basic idea of requiring four boundaries to search the center of an object may limit their performance ceiling. 

\begin{figure*}[t]
  \centering
  \includegraphics[width=0.95\linewidth]{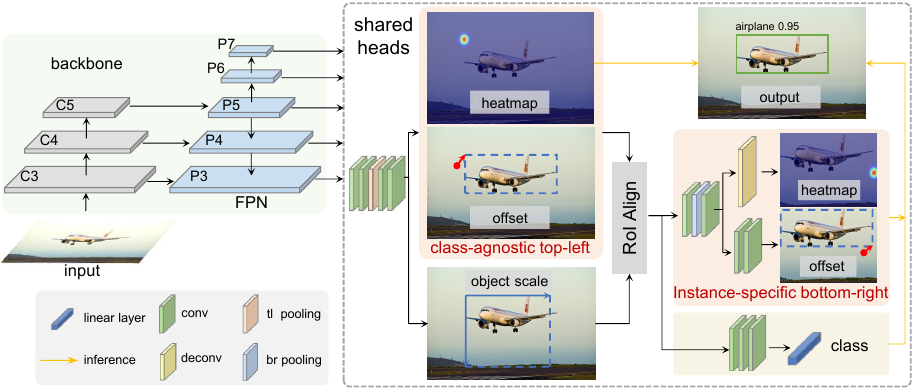}
  \caption{The framework of the proposed Corner2Net. Corner2Net has two cascade stages to conquer the corner-based detection task. In the first stage, all class-agnostic top-left corners are located on the image-level heatmap and adjusted by offsets, and each RoI space is determined to establish an association between two cascade corners. Next, the RoI features are fed into the second stage, where the precise bottom-right corner of each specific instance is obtained by its instance-level heatmap and offset. Meanwhile, the category is predicted by a lightweight head with rich instance semantics. }
  \label{fig:architecture}
\vspace{20pt}
\end{figure*}

\subsection{Corner-based Detectors}
The corner-based paradigm adopts a more bionic route, which evades the anchor generation mechanism and then constructs the bounding box of the instance by pairing corners. The concept is first introduced by CornerNet~\cite{law2018cornernet}, which predicts the top-left and bottom-right corners on heatmaps separately and matches them by measuring the distance of the corresponding instance embedding. To better perceive the visual patterns within proposals and suppress the wrong decoded boxes, CenterNet~\cite{duan2019centernet_duan} improves the correctness of the corner pair by upgrading them into keypoint triplets via an additional predicted center point. With the combination of position and geometry information, CentripetalNet~\cite{dong2020centripetalnet} devises a 2D centripetal shift to clearly judge whether the two corners are from the same instance. With greater progress, an affinity function~\cite{ijcai2022p203_CornerAffinity} is devised to boost the accuracy and box quality by incorporating both structure and context measurements. 
These works all contribute to optimizing the accuracy of the corner-matching process, yet it is still not perfect. Furthermore, the difficulty of learning two corners of the same category due to weak semantics is not considered. And they all adopt Hourglass as the backbone which is not friendly to follow. 
In this paper, we refresh the corner-based detection framework and propose Corner2Net to fully demonstrate the true power of this detection paradigm.

\section{Method}

\subsection{Framework of Cascade Corner Detection}
The corner-based detection paradigm is to determine the position of the object by finding its top-left and bottom-right corners. Existing methods utilize two parallel detection heads to predict two class-specific keypoints respectively, thus the post-processing is necessary to pair the independent corners into boxes. Different from the aforementioned pipeline, we divide the prediction task of the two corners into two steps instead of synchronously predicting them. As shown in Figure~\ref{fig:architecture}, the first step is to exclusively mine the top-left corner points of all instances regardless of category, and return the RoI space for each instance. Next, the second step is to extract the RoI features with rich context to predict the more precise instance-specific bottom-right corner by the single-channel heatmap. And the object category is also obtained in the second step with a simple classification head via instance semantics. Obviously, these two solving steps endow the proposed Corner2Net to be a two-stage detector. Next, we elaborate on each component for cascade corner detection. 

The previous corner-based detectors~\cite{law2018cornernet,duan2019centernet_duan,ijcai2022p203_CornerAffinity} are all developed based on the keypoint-friendly hourglass-104~\cite{newell2016stacked_hourglass} backbone to gain the best performance. But it has unpleasant training costs and inference speed. Instead of that, to align with commonly used strategies in other detectors, we also exploit the structure of multi-level feature prediction. Given an image with ${H\times W}$ pixels, it is input into the backbone to gain high-level semantics. The deep feature maps C3 to C5 are extracted and fed into FPN. The feature levels P3 to P7 of FPN are used for subsequent prediction through the shared heads. The downsampling ratios $s \in \left( 8, 16, 32, 64, 128 \right)$ lead to the output maps with the size of $\frac{H}{s} \times \frac{W}{s}$. In addition, we assign objects of different sizes to FPN layers based on the previous strategy~\cite{tian2019fcos,zhou2021probabilistic_centernet2}. Thus, Corner2Net uses multi-level features to help predict the corners of multi-scale objects. Corner2Net can flexibly connect common backbones (\textit{e.g.}, ResNet~\cite{he2016deep_resnet}, and ResNeXt~\cite{xie2017aggregated_ResNeXt}), and experimental results prove that robust accuracy can be obtained.

\subsection{Mine Every Class-agnostic Top-left Corner}
In the proposed cascade corner detection framework, the main goal of the first stage is to fully explore the top-left corner keypoints of all instances. We decouple position and category prediction and only estimate the class-agnostic keypoints for the top-left corner. Specifically, at each FPN feature level, Corner2Net predicts a $\frac{H}{s} \times \frac{W}{s}$ heatmap with a single channel via several convolution layers to estimate all top-left corners of the cared objects at the current level. We also adopt the corner pooling~\cite{law2018cornernet} to encode more visual evidence for the corners. For each top-left corner, the only ground-truth pixel on the heatmap is positive and set to 1 and other pixels are negative and set to 0. During training, the corner target is mapped as a Gaussian region~\cite{law2018cornernet}, reducing the penalty for the pixels near the ground-truth corners and helping balance positive and negative samples. We employ the class-agnostic focal loss~\cite{lin2017focalloss} with the distance-aware penalty as the optimization target of the top-left corner heatmap, \textit{i.e.}, 
\begin{equation}
\resizebox{0.9\linewidth}{!}{$
\mathcal{L}_{h}^{tl} \!= \!-\frac{1}{N_p}\sum\limits_{x=1}^{W^{\prime}}\sum\limits_{y=1}^{H^{\prime}}
\begin{cases}
(1\!-\!\acute{{p_{xy}}})^\alpha  \log \left(\acute{{p_{xy}}}\right), \! &\! \text{if $p_{xy}$ \!= \!1} \\
\left(\!1\!-\!p_{xy}\!\right)^\beta  \acute{{(p_{xy})}}^\alpha  \log(1-\acute{{p_{xy}}}), \! & \! ~\text{\!\!\!otherwise}
\end{cases}$}
\label{equation2}
\end{equation}
where $N_p$ denotes the sum of objects. $W^{\prime}$ and $H^{\prime}$ are the width and height of the heatmap. $p_{xy}$ is the ground-truth value at the coordinate $(x,y)$ on the top-left heatmap and $\acute{{p_{xy}}}$ is the predicted value in the same location. The hyper-parameter $\alpha$ can adjust the weights for hard and easy samples. The $\beta$ is used to control the term of distance-aware penalty. Referring to previous work~\cite{law2018cornernet}, we set $\alpha$ and $\beta$ to 2 and 4. 

After getting the multi-level predicted heatmaps, we need to extract keypoints with higher scores and map the coordinates back to the original image resolution. Due to the existence of the downsampling ratio at each layer of FPN, there is a discretization error for the coordinates on the downsampled heatmap. To compensate for this lost precision, Corner2Net predicts a position offset map for each top-left heatmap at different FPN levels, as shown in Figure~\ref{fig:architecture}. We adopt the Smooth L1 Loss~\cite{girshick2015fastrcnn_smoothl1} as the objective of the offset map, \textit{i.e.},
\begin{small}
\begin{equation}
    \mathcal{L}_{o}^{tl} \!=\! \frac{1}{N_p}\sum_{j=1}^{N_p} \mathrm{SmoothL1Loss}( (\Delta x_j^{\prime}, \Delta y_j^{\prime}), (\Delta x_j, \Delta y_j))
\end{equation}
\end{small}
where $ (\Delta x_j^{\prime}, \Delta y_j^{\prime})$ are the predicted values on the output offset map. $(\Delta x_j, \Delta y_j)$ are the corresponding ground truth and can be calculated by $(x_j-x_j^f,y_j-y_j^f)$. Here, $(x_j, y_j)$ is the discretized position on the heatmap and $(x_j^f, y_j^f)$ is the float coordinate of the ground-truth corner. $N_p$ means the number of positives. Thus, the extracted top-left corner location on the heatmap can be refined by the predicted offset. It is worth mentioning that this offset compensation is of great significance to the detection for small objects. 

\subsection{Establish Association between Corners via the RoI Space}
To achieve cascade corner detection, we intend to pinpoint the specific bottom-right keypoint for each top-left corner via its region of interest. This means that we need to determine the RoI space for establishing an association between the top-left and bottom-right corners in the first stage. In each FPN layer, Corner2Net also outputs a regression map that indicates an instance scale at each top-left positive position. For training, we utilize the GIoU loss~\cite{rezatofighi2019generalized_GIoU} as the objective function, which is formulated as: 
\begin{equation}
\mathcal{L}_{b} = \frac{1}{N_p}\! \sum_{j=1}^{N_p} \mathcal{L}_{GIoU}\left[(z_j, z_j), (z_j^f, z_j^f)\right]
\end{equation}
where $z_j$ and $z_j^f$ are the predicted scale and the ground-truth value, respectively. The ground-truth scale is the maximum value of the object's width/height. At this point, we can decode a proposal box $\left(x_{tl}, y_{tl}, z, s_{tl}\right)$ at the top-left corner location. Here, $\left(x_{tl}, y_{tl}\right)$ is the top-left corner coordinate refined by its offset and $s_{tl}$ is the confidence predicted on the heatmap. The width and height of the proposal are equal to the approximate scale $z$, that is, the expected RoI space is square. The square RoI space can help minimize the axial compression ratio, so that the feature blocks obtained by RoIAlign~\cite{he2017mask_rcnn} can retain more guidance of spatial information.

Note that the proposal RoI space is provided via a rough predicted object scale, which cannot completely enclose the instance. Accordingly, to ensure that the bottom-right corner can fall inside the proposal RoI, we expand the vanilla RoI space by enlarging its regressed scale, and related formulations are described as follows: 
\begin{equation}
    \left(x_{tl}, y_{tl}, z_e, s_{tl}\right) {\longleftarrow} \left(x_{tl}, y_{tl}, (1 + \eta ) \cdot z, s_{tl}\right)
\label{equation:enlarge}
\end{equation}
where $z_e$ is the enlarged scale and $\eta$ denotes the enlarge factor. Then, Corner2Net can locate the bottom-right keypoint of each instance within the specific enlarged RoI space via the next stage. 

\subsection{Locate Each Instance-specific Bottom-right Corner}
In the second stage of the cascade corner detection framework, the bottom-right corner is predicted for the specific instance in the RoI space determined by the corresponding top-left corner, as shown in Figure~\ref{fig:architecture}. Clearly, our solution devises two cascade corners (or an associated corner pair) instead of two completely independent corners, avoiding the heuristic corner-matching algorithms that are indispensable in parallel corner-based pipelines~\cite{law2018cornernet,duan2019centernet_duan,ijcai2022p203_CornerAffinity}. 

We use the RoIAlign~\cite{he2017mask_rcnn} operation with an output size of $14\times14$ to extract the features used to locate bottom-right corners. Then, there are two convolutions with a bottom-right corner pooling layer in between. Next, a deconvolution layer is placed behind to output a single-channel heatmap with the $28\times28$ pixels. For training, similarly to the top-left point, we render the unique ground-truth bottom-right corner of each object as a Gaussian region~\cite{law2018cornernet} by $\operatorname{exp}(-\frac{x_k^2+y_k^2}{2\sigma^2})$, and the radius $\sigma$ is calculated by the object size. After that, we need to project each point $(x_{k}, y_{k})$ in the RoI region of the original image space onto the coordinate system of the heatmap branch. The transformation can be formulated as:
\begin{equation}
\begin{cases}
    x_h = (x_{k} - x_{tl}) \cdot \frac{m}{z_e}
    \\
    y_h = (y_{k} - y_{tl}) \cdot \frac{m}{z_e}
\end{cases}
\label{equation:transform}
\end{equation}
where $(x_h, y_h)$ is the coordinate on the heatmap. $m$ denotes the side length of the heatmap and the default value is 28. $x_{tl}$, $y_{tl}$, and $z_e$ are the same as in Eq.~\ref{equation:enlarge}. And we employ the binary cross entropy loss to train the instance-specific bottom-right heatmap branch, and the function is as follows: 
\begin{small}
\begin{equation}
\begin{aligned}
    \mathcal{L}_{h}^{br} \!= \!\frac{1}{N_{fb} \! \cdot\! m^2} \sum_{n=1}^{N_{fb}} \sum_{x=1}^{m} \sum_{y=1}^{m}\! -& p_{nxy}\cdot \operatorname{log}\left(\phi(\acute{p_{nxy}})\right) \\
    - & \left(1\!-\!p_{nxy}\right) \! \cdot\! \operatorname{log} \left( 1\!-\!\phi(\acute{p_{nxy}}) \right)
\end{aligned}
\end{equation}
\end{small}
where $p_{nxy}$ and $\acute{p_{nxy}}$ are the ground truth and predicted value, respectively. $\phi (\cdot)$ denotes the sigmoid function. $N_{fb}$ is the number of foreground RoIs. For inference, the predicted heatmap first undergoes sigmoid function and $3\times3$ max pooling, and then the top-1 keypoint is picked to decode the bottom-right corner location. 

It is easy to find that, as illustrated in Eq.~\ref{equation:transform}, there is a quantization error in the instance-specific heatmap. Therefore, it is necessary to return an offset to compensate for the precision loss. Specifically, along with the heatmap branch, we apply two convolution layers to predict an offset for the specific bottom-right corner, so as to fine-tune the position of the bottom-right keypoint. During training, the Smooth L1 Loss~\cite{girshick2015fastrcnn_smoothl1} is chosen as the objective function:  
\begin{small}
\begin{equation}
    \mathcal{L}_{o}^{br}\! =\! \frac{1}{N_b}\sum_{j=1}^{N_b} \mathrm{SmoothL1Loss}( (\Delta u_j^{\prime}, \Delta v_j^{\prime}), (\Delta u_j, \Delta v_j))
\label{equation:br_off_loss}
\end{equation}
\end{small}
where $N_b$ is the number of RoI spaces. $\Delta u_j^{\prime}$ $\Delta v_j^{\prime}$ are the predicted offsets in the horizontal and vertical directions. During inference, the coordinates of the bottom-right corner $(x_{br}, y_{br})$ on the input image can be calculated by Eq.~\ref{equation:br_decode}. 
\begin{equation}
\begin{cases}
    x_{br} = \left( x_{br}^\prime + \Delta u_j^{\prime} \right) \cdot \frac{z_e}{m} + x_{tl}
    \\
    y_{br} = \left( y_{br}^\prime + \Delta v_j^{\prime} \right) \cdot \frac{z_e}{m} + y_{tl}
\end{cases}
\label{equation:br_decode}
\end{equation}
where $(x_{br}^\prime, y_{br}^\prime)$ is the keypoint extracted on the bottom-right heatmap. $z_e$ denotes the enlarged scale of RoI space and $m$ is the same as in Eq.~\ref{equation:transform}. Finally, we obtain the purified bottom-right corner and can utilize this associated corner pair to decode a high-quality detection box. 

\subsection{Predict Category by RoI Features with Rich Instance Semantics}
In the proposed Corner2Net, the classification score can be obtained by a lightweight head in the second stage. Compared to those existing corner-based detectors that determine the object category through the channel id of unrobust class-specific keypoints, Corner2Net utilizes several convolutions and two fully connected layers to perform classification by RoI features which preserve richer instance semantics. To alleviate the interference of background noise on the classifier, we utilize vanilla RoIs without being enlarged to produce the features with the spatial size of $7 \times 7$ to be classified. And we adopt the cross entropy~\cite{ren2015fasterrcnn} loss function to train the classification head. During inference, we take into account both the localization score and the classification score. Note that the score of the cascade corner pair represents the localization score of the constructed detection box. Then, the score of the final detection box is computed as follows:
\begin{equation}
    s_{box} = \sqrt{\left[0.5\left(s_{tl}+s_{br}\right)\right]_{loc} \cdot s_{cls}}
\end{equation}
where $s_{cls}$ is the predicted class confidence. $\left[ \cdot \right]_{loc}$ represents the localization term, which is the arithmetic mean of the top-left and bottom-right corner confidence. 

\section{Experiments}
\subsection{Datasets and Evaluation Metrics}
\paragraph{MS-COCO.}
MS-COCO~\cite{lin2014microsoft_coco_dataset} is a large-scale detection dataset with a resolution of about $600\times800$ and labels containing 80 categories. Our Corner2Net is trained on the train2017 set including $118k$ images. We report the performance on the test-dev set including $20k$ images for the state-of-the-art comparison with other published detectors. We perform ablation studies on the COCO val2017 set including $36k$ instances within $5k$ images. 
\paragraph{CityPersons.}
We perform evaluations on CityPersons~\cite{cordts2016cityscapes} dataset to verify the robustness of our model in the scene with dense occlusion. We filter the vanilla dataset, and merge the main annotations of pedestrians and riders. As a result, the training set contains 2471 images and 18k persons. The evaluation results are achieved on the validation set including 439 images and 3666 persons.
\paragraph{UCAS-AOD.}
UCAS-AOD~\cite{zhu2015orientation} is selected to demonstrate the generalization and applicability of the proposed Corner2Net when facing many similar and evenly symmetrically distributed targets. There are 1000 high-resolution aerial images in this dataset. And the total number of annotated aircraft is 7482.
\paragraph{Evaluation metrics.} We adopt commonly used AP (Average Precision) and FPS (frames per second) as metrics to measure detection accuracy and inference speed, respectively.

\subsection{Implementation Details}
During training, we adopt the SGD optimizer to minimize the training loss and we initialize the backbone using parameters pretrained on the ImageNet~\cite{krizhevsky2017imagenet}. And our model is trained on 8 RTX 3090 GPUs with two images on each for 24 epochs. The initial learning rate is set to 0.02 for the first 16 epochs, and it is decayed by $\times 10$ at the 16th epoch and 22nd epoch.

During inference, the predicted class-agnostic top-left heatmap is performed $3\times3$ max pooling, and then the top 128 proposal keypoints are extracted to search each specific bottom-right corner. The duplicate boxes are filtered out by NMS with an IoU threshold of 0.6. For the single-scale testing, each image is resized to a fixed short side of 640 pixels. When performing multi-scale testing, the short side of each resized image is in the range of $\left[ 400:200:1400\right]$.

\begin{table*}[t]
	\caption{Comparisons with state-of-the-art methods in term of accuracy ($\%$) on the MS-COCO test-dev set. $(*)$ means the implementation of multi-scale testing. The full name of ``DCN'' is the Deformable Convolutional Networks. ``T'' means ``tiny''.}
	\begin{center}
		\setlength{\tabcolsep}{1.5mm}
  {
			\small
			\begin{tabular}{l|l|c|c|c|ccc|ccc}
				\hline  
    
				\hline
				Method \textbf{(test-dev set)} & Backbone & FPS & Params (M) & Epoch & $\mathrm{AP}$ & $\mathrm{AP_{50}}$ & $\mathrm{AP_{75}}$ & $\mathrm{AP_{S}}$ & $\mathrm{AP_{M}}$ & $\mathrm{AP_{L}}$ \\
				\hline 
				\textbf{Center-based:} & & & & & & & & & & \\
				Faster R-CNN~\cite{ren2015fasterrcnn} & ResNet-101 & 9.5 & 60.75 & 24 & 36.2 & 59.1 & 39.0 & 18.2 & 39.0 & 48.2  \\

                Cascade R-CNN~\cite{cai2018cascadercnn} & ResNet-101 & 8.1 & 88.39 & 24 & 42.8 & 62.1 & 46.3 & 23.7 & 45.5 & 55.2 \\
				
			    RetinaNet~\cite{lin2017focalloss} & ResNet-101 & 8.9 & 56.96 & 36 & 40.8 & 61.1 & 44.1 & 24.1 & 44.2 & 51.2 \\

                Grid R-CNN~\cite{lu2019grid_rcnn} & ResNet-101 & 8.0 & 83.31 & 24 & 41.5 & 60.9 & 44.5 & 23.3 & 44.9 & 53.1 \\

                Libra R-CNN~\cite{pang2019libra_rcnn} & ResNet-101 & 8.8 & 60.78 & 24 & 41.1 & 62.1 & 44.7 & 23.4 & 43.7 & 52.5 \\

                Reppoints~\cite{yang2019reppointsv1} & ResNet-101 & 8.4 & 55.62 & 24 & 41.0 & 62.9 & 44.3 & 23.6 & 44.1 & 51.7 \\

                CenterNet~\cite{zhou2019centernet_zhou}  &Hourglass-104 & 4.2 & 186.44 & 200 & 42.1 & 61.1 & 45.9 & 24.1 & 45.5 & 52.8 \\

				FCOS~\cite{tian2019fcos} & ResNeXt-101-DCN & 6.9 & 89.79 & 24  & 46.6 & 65.9 & 50.8 & 28.6 & 49.1 & 58.6 \\
                SAPD~\cite{zhu2020soft_SAPD} & ResNeXt-101-DCN & 6.2 & 101.20 & 24 & 46.6 & \textbf{66.6} & 50.0 & 27.3 & 49.7 & \textbf{60.7}  \\
				\hline
				\textbf{Parallel corner-based:} & & & & & & & & & &  \\
				CornerNet~\cite{law2018cornernet} & Hourglass-104 & 3.9 & 192.04 & 200 & 40.6 & 56.4 & 43.2 & 19.1 & 42.8 & 54.3 \\
				CenterNet~\cite{duan2019centernet_duan} & Hourglass-104 & 3.3 & 201.20 & 190 & 44.9 & 62.4 & 48.1 & 25.6 & 47.4 & 57.4 \\		
                CentripetalNet~\cite{dong2020centripetalnet} & Hourglass-104 & 3.4 & 197.76 & 210 & 45.8 & 63.0 & 49.3 & 25.0 & 48.2 & 58.7 \\

                CornerAffinity~\cite{ijcai2022p203_CornerAffinity} & Hourglass-104 & 3.7 & 194.60 & 320 & 46.3 & 64.0 & 49.9 & 27.4 & 49.3 & 58.7 \\
		
				\textbf{Cascade corner-based:} & & & & & & & & \\
                \rowcolor{gray!10}
				Corner2Net (Ours) & ResNeXt-101-DCN & {8.0} & {119.23} & \textbf{24} & \textbf{46.8} & 65.7 & \textbf{51.9} & \textbf{29.3} & \textbf{49.8} & 57.9 \\
                \rowcolor{gray!10}
				Corner2Net$^*$ (Ours) & ResNeXt-101-DCN & - & - & 24 & 47.8 & 65.5 & 53.9 & 31.5 & 50.2 & 57.6 \\
  
                \rowcolor{gray!10}
				Corner2Net (Ours) & ResNet-101 & {9.4} & {67.87} & 24 & 43.8 & 61.8 & 48.5 & 26.5 & 46.7 & 53.7 \\
                \rowcolor{gray!10}
                Corner2Net$^*$ (Ours) & ResNet-101 & - & - & 24 & 45.0 & 61.8 & 50.6 & 29.5 & 47.3 & 53.8 \\
                \rowcolor{gray!10}
				Corner2Net (Ours) & SwinTransformer-T & \textbf{11.2} & \textbf{53.03} & {24} & 46.0 & 65.2 & 50.7 & 28.2 & 48.4 & 57.9 \\
                \rowcolor{gray!10}
				Corner2Net$^*$ (Ours) & SwinTransformer-T & - & - & 24 & 46.6 & 63.9 & 52.5 & 30.2 & 49.0 & 57.8 \\

        \hline

        \hline

		\end{tabular}}
  
	\end{center}

	\label{table:sota}
    \end{table*}

\begin{table*}[t]
\caption{Comparisons in terms of box quality on MS-COCO val2017 set. Higher IoU corresponds to higher-quality detection boxes. ``-'' denotes the model weights of the original paper are not available.}
	\begin{center}
		\setlength{\tabcolsep}{1.5mm}
  {
		
			\small
			\begin{tabular}{l|l|c|ccc|ccc}
        \hline

        \hline
                Method \textbf{(val set)} & Backbone & Epoch & $\mathrm{AP}$ & $\mathrm{AP_{50}}$ & $\mathrm{AP_{60}}$ & $\mathrm{AP_{70}}$ & $\mathrm{AP_{80}}$ & $\mathrm{AP_{90}}$ \\
        \hline
                CornerNet~\cite{law2018cornernet} & Hourglass-104 & 200 & 40.6 & 56.1 & 52.0 & 46.8 & 38.8 & 23.4 \\
                CenterNet~\cite{duan2019centernet_duan} & Hourglass-104 & 190 & 44.8 & 62.5 & 58.1 & 52.3 & 42.9 & 22.9 \\		
                CentripetalNet~\cite{dong2020centripetalnet} & Hourglass-104 & 210 & 44.7 & 62.6 & 58.1 & 52.8 & 42.3 & 22.3 \\
                CornerAffinity~\cite{ijcai2022p203_CornerAffinity} & Hourglass-104 & 320 & 45.1 & 62.9 & - & -& - & - \\

                \rowcolor{gray!10}
				Corner2Net  (Ours) & ResNeXt-101-DCN & \textbf{24} & \textbf{46.3} & \textbf{65.1} & \textbf{61.5} & \textbf{55.6} & \textbf{44.6} & 22.4 \\
        \hline

        \hline
		\end{tabular}}
  
	\end{center}
        
	\label{table:coco_val}
    \end{table*}

\begin{table}[t]
        \caption{Accuracy comparisons on CityPersons ($\mathrm{AP_{50}^c}$) and UCAS-AOD ($\mathrm{AP^{u}_{50}}$) dataset. SwinT-T means SwinTransformer (tiny).}
	\begin{center}
		\setlength{\tabcolsep}{1.5mm}{
			\small
			\begin{tabular}{l|c|c|c|c}
				\hline  

				\hline
			
				Model & Backbone & epoch & $\mathrm{AP_{50}^c}$ & $\mathrm{AP^{u}_{50}}$ \\
				\hline
                CornerNet & Hourglass-104 & 50 & 29.1 & 79.1 \\
                CenterNet & Hourglass-104 & 50 & 51.5 & 86.8 \\
                CentripetalNet & Hourglass-104 & 50 & 54.7 & 93.6 \\
                CornerAffinity & Hourglass-104 & 50 & 64.9 & 96.3 \\       
                \rowcolor{gray!10}
                Corner2Net (Ours) & SwinT-T & 12 & 65.3 & 97.1 \\

				\hline

                \hline
		\end{tabular}}
		
	\end{center}
	\label{table:ucas_city}
    \end{table}

\subsection{Comparison with State-of-the-art Methods}
\paragraph{Main results on COCO.}
To demonstrate the performance of the proposed Corner2Net, we conduct the accuracy comparison with state-of-the-art detectors on the COCO test-dev, and the evaluation results are reported in Table~\ref{table:sota}. Generally, the proposed Corner2Net equipped with the backbone ResNeXt-101-DCN-32$\times$8d can achieve the $\mathrm{AP}$ of 46.8\%/47.8\% under the single-scale/multi-scale testing, outperforming the counterparts with the same settings. 
\paragraph{Corner2Net versus parallel corner-based detectors.}
For corner-based detection methods, which is the paradigm to which our method belongs, our method showcases state-of-the-art accuracy and requires only a small number of training epochs. Specifically, in comparison with CornerAffinity~\cite{ijcai2022p203_CornerAffinity} that enjoys an almost ceiling-level corner-matching algorithm, our Corner2Net brings the $+$1.7\%$\mathrm{AP_{50}}$ improvement using 7.5\% epochs (24 \textit{vs.} 320). This is because Corner2Net avoids struggling in matching unrobust class-specific corners via reliable cascade class-agnostic corners. 
\paragraph{Corner2Net versus center-based detectors.}
Benefiting from decoupled instance-informative-rich classification and matching-free corner detection, Corner2Net surpasses the common center-based FCOS~\cite{tian2019fcos} (\textit{i.e.}, 46.8\% \textit{vs.} 46.6\% on $\mathrm{AP}$) and SAPD~\cite{zhu2020soft_SAPD} under the same settings. Moreover, compared to the performance from popular center-based detectors, \textit{e.g.}, Cascade R-CNN~\cite{cai2018cascadercnn} and Reppoints~\cite{yang2019reppointsv1}, Corner2Net with the same ResNet-101 backbone also lifts $\mathrm{AP}$ by at least 1.0\%. This phenomenon shows that searching for corner points is easier than center points when detecting objects, which supports the opinion stated in Section~\ref{section:intro}. 
Notably, compared to Grid R-CNN~\cite{lu2019grid_rcnn} using multiple keypoints to localize centers in the second stage, the $\mathrm{AP}$ of Corner2Net increases by 2.3\%, which demonstrates the superiority of the proposed framework that directly find two precise corner kepoints in two stages. 

\paragraph{Training efficiency.}
Besides, compared to other corner detectors that rely on hourglass networks and require 200/300 training epochs, Corner2Net with SwinTransformer (tiny)~\cite{liu2021SwinTransformer} backbone only needs 24 epochs to yield a pleasing $\mathrm{AP}$ of 46.0\%, surpassing the CornerNet~\cite{law2018cornernet} baseline by 5.4\% $\mathrm{AP}$. This shows that our corner detection framework is more efficient and convenient to explore further.  
\paragraph{Quality of the detection box.}
We hold the view that corner-based detectors can generate higher-quality boxes than center-based ones. To examine this, we conduct the accuracy evaluation at high IoU thresholds on MS-COCO val2017. As shown in Table~\ref{table:coco_val}, using only 12\% of the training epochs of CornerNet, our model comprehensively outperforms CornerNet in several indicators, including $\mathrm{AP_{50}}$, $\mathrm{AP_{60}}$, $\mathrm{AP_{70}}$, and $\mathrm{AP_{80}}$. 
It is worth noting that our Corner2Net reaches an impressive $\mathrm{AP_{80}}$ of 44.6\%, surpassing CornerNet by a large margin of 5.8\%. The 22.4\% $\mathrm{AP_{90}}$ of Corner2Net has a 1\% gap compared to CornerNet. This is because the downsampling stride of the feature map in Corner2Net is much larger than that of the vanilla CornerNet, making the compensation of quantization error more difficult. Nonetheless, these results are sufficient to demonstrate that the proposed Corner2Net has achieved a reasonable trade-off between box quality and computational costs.

\subsection{Comparisons on Inference Speed}
The proposed Corner2Net no longer relies on the sluggish hourglass backbone network, and matching-free cascade corner pairs facilitate a faster inference speed. We test the inference speed of all models on Titan Xp GPU. As shown in Table~\ref{table:sota}, CornerNet, CenterNet~\cite{duan2019centernet_duan}, CentripetalNet~\cite{dong2020centripetalnet}, CornerAffinity~\cite{ijcai2022p203_CornerAffinity} and our Corner2Net yield 3.9, 3.3, 3.4, 3.7 and 8.0 FPS, respectively. Further, our Corner2Net with SwinTransformer (tiny) is 2.9 times faster than vanilla CornerNet with Hourglass-104 and lifts $\mathrm{AP}$ by 5.4\%. 
During inference, CornerNet~\cite{law2018cornernet} processes 100$^2$ proposals constructed by 100 top-left and 100 bottom-right corners, while our Corner2Net only extracts 128 cascade corner pairs to achieve the promising performance. 

\subsection{Comparisons in Extreme Scenarios}
To further demonstrate the practicality and generalization, we evaluate the proposed Corner2Net in some extreme scenes, including the CityPersons dataset with dense occlusion and the UCAS-AOD dataset containing the symmetrical arrangement of numerous rotated similar objects. The results are reported in Table~\ref{table:ucas_city}. These parallel corner-based methods suffer from performance bottlenecks due to heuristic corner-matching algorithms. Compared to these corner baselines, the proposed cascade corner-based Corner2Net achieves the best $\mathrm{AP_{50}^c}$ of 65.3\% and $\mathrm{AP^{u}_{50}}$ of 97.1\%. This shows that our method of predicting the cascade associated corner pairs is more robust to extreme scenarios and the corner-matching free manner avoids confusion when matching corners of similar objects.

\begin{table}[t]
        \caption{Effect of the number ($k$) of extracted cascade corner pairs. The results are obtained on MS-COCO val2017 under the same backbone SwinTransformer (tiny). }
	\begin{center}
		\setlength{\tabcolsep}{1.5mm}{
			\small
     
			\begin{tabular}{l|cccccc}
				\hline  
    
				\hline

				top-k & $\mathrm{AP}$ & $\mathrm{AP_{50}}$ & $\mathrm{AP_{75}}$ & $\mathrm{AP_{S}}$ & $\mathrm{AP_{M}}$ & $\mathrm{AP_{L}}$ \\
				\hline
                $k=$100 & 45.6 & 64.5 & 50.0 & 29.8 & 48.7 & 59.7 \\
                $k=$128 & 45.7 & 64.7 & 50.1 & 29.9 & 48.8 & 59.8 \\
                $k=$256 & 45.7 & 64.6 & 50.3 & 30.1 & 48.9 & 59.7 \\
                $k=$512 & 45.6 & 64.5 & 50.3 & 30.1 & 48.9 & 59.7 \\
                $k=$1024 & 45.6 & 64.3 & 50.3 & 30.0 & 48.8 & 59.7 \\
				\hline

                \hline
		\end{tabular}}
	\end{center}

	\label{table:ablation_topk}
    \end{table}

\begin{figure*}[!t]
  \centering
  \includegraphics[width=\linewidth]{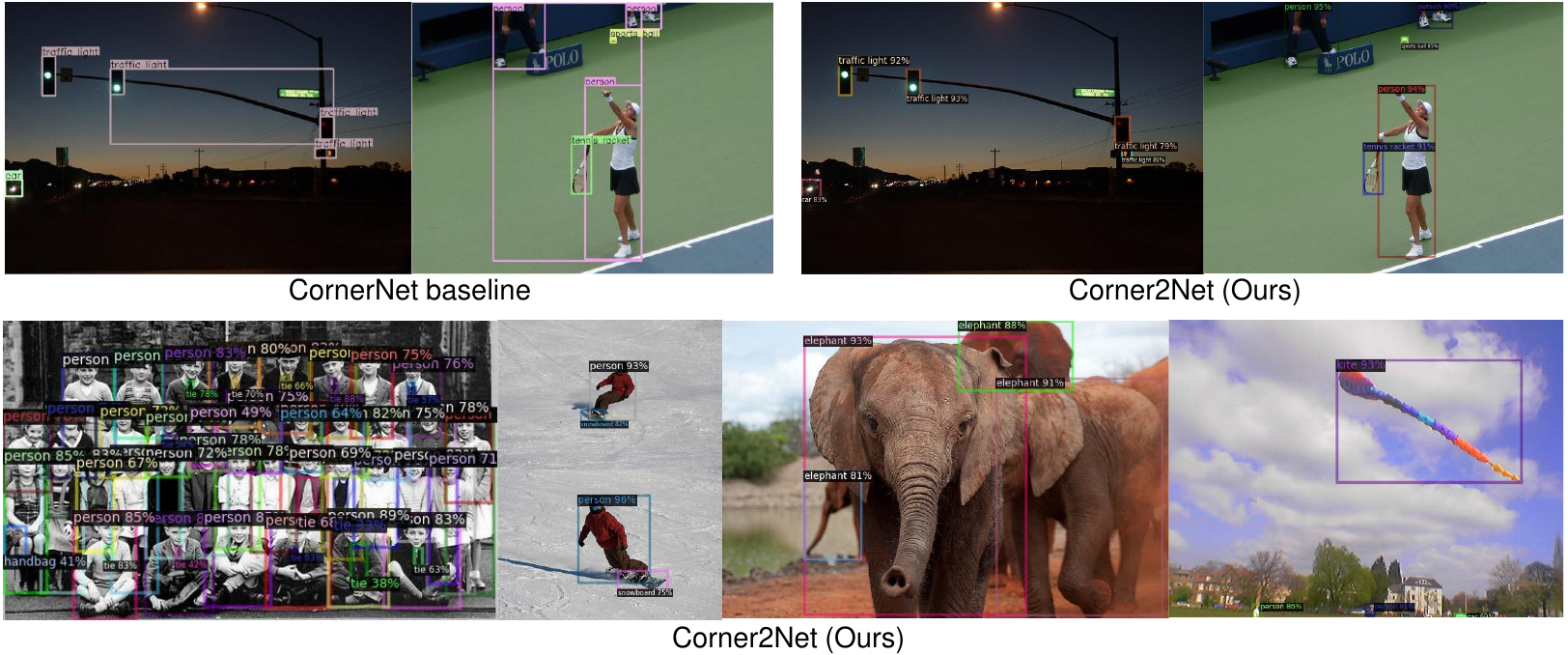}
  \caption{Qualitative detection results of CornerNet baseline and the proposed Corner2Net on MS-COCO val2017 set.}
  \label{fig:vis}
\vspace{30pt}
\end{figure*}

\subsection{Ablation Study}
In this section, we implement ablation analysis to verify the effectiveness of each component and explore better hyper-parameters. And the results are obtained on the MS-COCO val2017 set.
\paragraph{Effect of the number of cascade corner pairs.} 
In the inference stage, the number of cascade corner pairs is equal to the number of top-left keypoints extracted on the heatmap. As shown in Table~\ref{table:ablation_topk}, the accuracy of Corner2Net is not sensitive to the number of corner pairs, proving that it is robust and sufficient to locate objects by predicting class-agnostic cascade corners. We choose the value of $k$ to be 128.

\begin{table}[t]
        \caption{Effect of the enlarge factor ($\eta$) in Eq.~\ref{equation:enlarge}. The results are obtained on MS-COCO val2017 under the same backbone SwinTransformer (tiny). }
	\begin{center}
		\setlength{\tabcolsep}{1.5mm}{
			\small
			\begin{tabular}{c|cccccc}
				\hline  
    
				\hline

				enlarge factor & $\mathrm{AP}$ & $\mathrm{AP_{50}}$ & $\mathrm{AP_{75}}$ & $\mathrm{AP_{S}}$ & $\mathrm{AP_{M}}$ & $\mathrm{AP_{L}}$\\
				\hline
                $\eta=$0.00 & 32.5 & 52.5 & 33.5 & 18.3 & 31.2 & 49.7 \\
                $\eta=$0.10 & 45.0 & 64.0 & 49.1 & 28.4 & 48.4 & 59.8 \\
                $\eta=$0.25 & 45.7 & 64.7 & 50.1 & 29.9 & 48.8 & 59.8 \\
                $\eta=$0.35 & 44.8 & 64.1 & 48.7 & 28.9 & 48.2 & 59.3 \\
                $\eta=$0.50 & 42.7 & 62.2 & 46.2 & 26.5 & 46.2 & 57.3 \\
				\hline

                \hline
		\end{tabular}}
		
	\end{center}

	\label{table:ablation_enlarge}
    \end{table}

\paragraph{Effect of the enlarge factor.}
We conduct experiments to find the optimal enlarge factor.
As shown in Table~\ref{table:ablation_enlarge}, the accuracy with $\eta=$0.10 results in an $\mathrm{AP}$ drop of 0.7\%, because some bottom-right corners fall outside the RoI space due to the smaller enlarge factor. A larger factor ($\eta=$0.5) also leads to a 3\% decrease on $\mathrm{AP}$, because the larger RoI space contains more noise, which is not conducive to locating the instance-specific bottom-right corner. Thus, we set $\eta$ to 0.25 in all experiments.

\subsection{Qualitative Analysis}
\begin{figure}[!t]
  \centering
  \includegraphics[width=\linewidth]{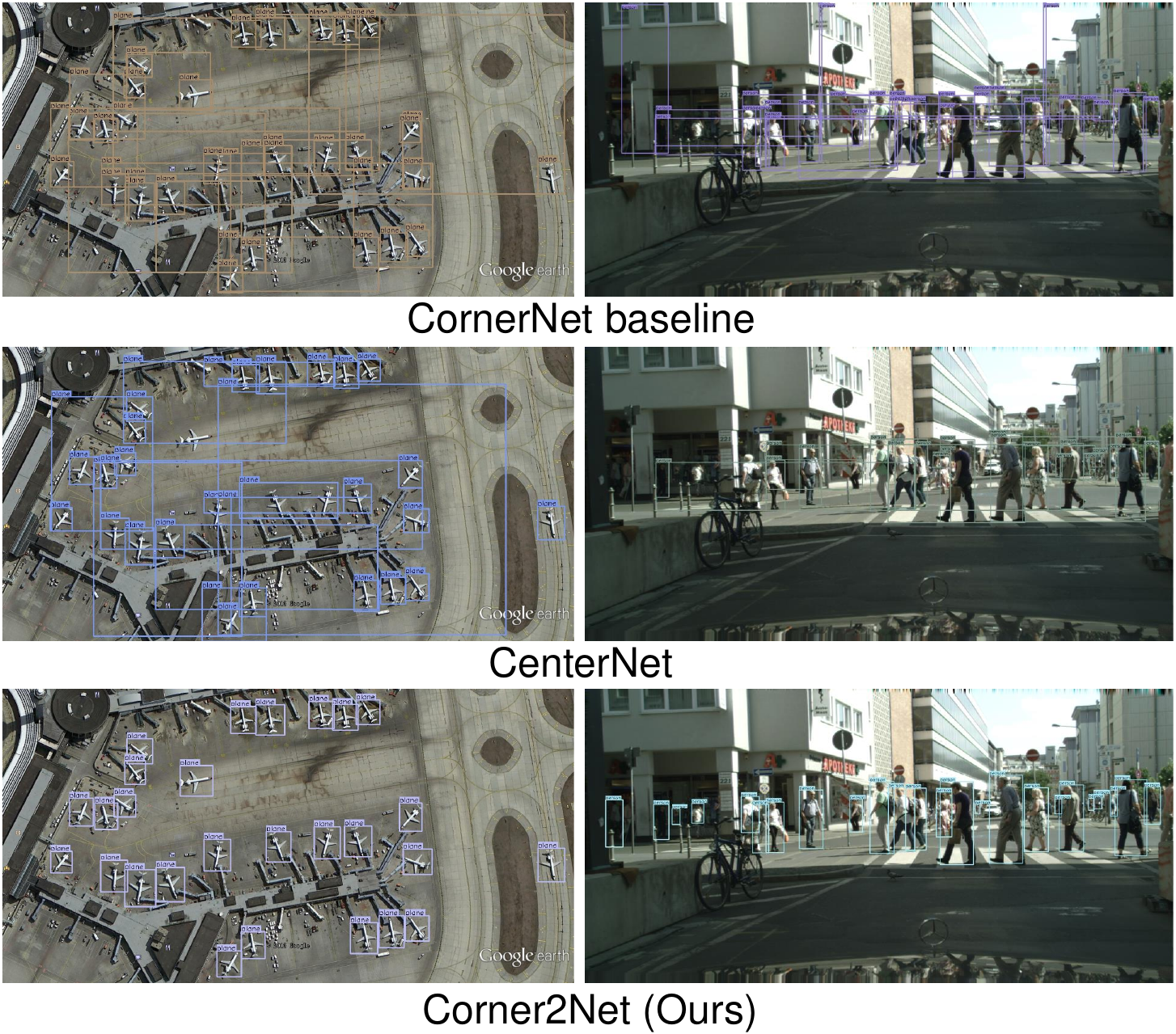}
  \caption{Visualization results on UCAS-AOD (left column) and CityPersons (right column) datasets.}
  \label{fig:vs3}
\vspace{30pt}
\end{figure}

\paragraph{COCO.}
As illustrated in Figure~\ref{fig:vis}, CornerNet baseline produces incorrect boxes because its corner-matching algorithm can not distinguish objects with similar appearances. Compared to it, the mismatched corners that often occur in previous corner-based methods do not exist in the results of the proposed Corner2Net due to the corner-matching-free manner. Furthermore, the high-quality visual detection boxes decoded by our Corner2Net have precise boundaries, which verifies the excellent $\mathrm{AP_{80}}$ and $\mathrm{AP_{90}}$. Hence, the detection box decoded by the proposed cascade corner pair is more reliable than that obtained by matching corners heuristically. 
\paragraph{UCAS-AOD and CityPersons.}
Some visualization results of extreme scenarios are shown in Figure~\ref{fig:vs3}. 
CornerNet baseline outputs many false boxes due to confusion about similar objects. CenterNet mismatches some corners because the center points of some other objects may fall within the determination area of the current two corners.
The proposed Corner2Net can produce accurate bounding boxes when similar objects co-occur or are partially occluded. This indicates that our model enjoys strong robustness and practicality.

\section{Conclusion}
In this paper, we deeply analyze factors limiting the development of parallel corner-based methods and propose a novel cascade corner detection framework to get rid of these constraints. The proposed Corner2Net runs in a corner-matching-free manner and it is also more robust to different popular backbones. Both in accuracy and speed, it surpasses all existing corner-based detectors and enjoys great untapped potential. 
We hope that this novel cascade corner detection baseline will attract more researchers to revitalize this paradigm with much room for improvement. 

\newpage

\section*{Acknowledgements}
This work was supported by National Science and Technology Major Project (No.2022ZD0119402), "Pioneer" and "Leading Goose" R\&D Program of Zhejiang (No.2024C01142), National Natural Science Foundation of China (No.U21B2013).



\bibliography{mybibfile}

\end{document}